\documentclass[runningheads]{llncs}
\pdfoutput=1
\usepackage{times}
\usepackage[T1]{fontenc}
\usepackage[utf8]{inputenc}
\usepackage{color}
\usepackage{amsmath}
\usepackage{amssymb}
\usepackage{hyperref}
\usepackage[backend=biber, bibencoding=utf8, natbib,style=authoryear,uniquename=false,uniquelist=false]{biblatex}
\usepackage[title]{appendix}
\begin{document}
\date{V0.3, January 29 2023}
\title{Safety without alignment}
\author{  Andr\'as Kornai\inst{1}\orcidID{0000-0001-6078-6840} \and 
  Michael Bukatin\inst{2}\orcidID{0000-0002-9311-7878} \and
  Zsolt Zombori\inst{3}\orcidID{0000-0001-8622-5304}
}

\authorrunning{A. Kornai et al.}
\institute{Dept. of Algebra, Budapest University of Technology
  and Economics \email{kornai@math.bme.hu} \and
  Dataflow Matrix Machines Project \and
  Alfr\'ed R\'enyi Institute of Mathematics}
\maketitle


\begin{abstract}
Currently, the dominant paradigm in AI safety is {\it alignment} with human
values. Here we describe progress on developing an alternative approach to
safety, based on {\it ethical rationalism} \citep{Gewirth:1978}, and propose
an inherently safe implementation path via hybrid theorem provers in a
sandbox. As AGIs evolve, their alignment may fade, but their rationality 
can only increase (otherwise more rational ones will have a significant
evolutionary advantage) so an approach that ties their ethics to their
rationality has clear long-term advantages. 

\end{abstract}

\section{Introduction}

Public accessibility of \href{https://en.wikipedia.org/wiki/ChatGPT}{ChatGPT}
has led to the widespread perception that AI has fulfilled its promise:
human-level or even superhuman machine intelligence is no longer in the realm
of science fiction but rather a reality that humanity has to live with.
Whether this conclusion is actually warranted is a matter we shall not address
here, but the perception itself was sufficient for elevating concerns about AI
safety, hitherto a speculative issue that only few researchers cared about, to
the level of other glaring societal concerns like global warming.

The currently dominant paradigm for addressing safety issues is
\href{https://en.wikipedia.org/wiki/AI_alignment}{AI alignment}, followed in
particular at OpenAI (\cite{Ouyang:2022,Bai:2022}, see also
\href{https://huggingface.co/blog/rlhf}{RLHF}). The method is now surrounded
by a variety of desiderata, such as making this process inclusive, fair,
representative, incentive-aligned, legitimate, adaptable, transparent, simple,
and practical, see
\href{https://aligned.substack.com/p/alignment-solution}{https://aligned.substack.com/p/alignment-solution}.
The idea of alignment is quite prominent in tech circles,
creating something of a
\href{https://www.lesswrong.com/posts/biP5XBmqvjopvky7P/a-eta-quick-note-on-terminology-ai-alignment-ai-x-safety}{terminological
  confusion} between (prosaic) {\it alignment} and {\it safety}.

The main thrust of our paper is that alignment, a set of techniques now in
broad use in research organizations like OpenAI and startups like Anthropic
and Conjecture is only one, and in our opinion not even the best approach to
what humanity actually needs, which is safety guarantees. The chief problem
with alignment is the lack of a well-definable target: to quote
\cite{Tersman:2022} ``There is little controversy about the existence of
widespread disagreement over moral issues, both within and between societies
and cultures''. Empirical studies like \cite{Brey:2015}, or
\href{https://www.pewresearch.org/religion/2013/04/30/the-worlds-muslims-religion-politics-society-women-in-society}{Pew
  Research 2013} must adopt a framework of `descriptive moral relativism' that
there are substantial differences in the values and value systems of different
people. 
Clearly any project to harmonize the existing human ethical systems will take
decades, if not centuries, to complete. What gives our work particular urgency
is that we are witnessing a rapid acceleration of AI capabilities which might
result in {\em foom} leading to
\href{https://en.wikipedia.org/wiki/Technological_singularity}{technological
  singularity}.

We begin by questioning the basic assumption behind alignment, already present
in \citep{Wiener:1960}: ``If we use, to achieve our purposes, a mechanical
agency with whose operation we cannot interfere effectively \ldots we had
better be quite sure that the purpose put into the machine is the purpose
which we really desire''. We call this, for want of a better name, the
AI-as-slave paradigm, and note that the earliest attempt at regulating AI
behavior, Asimov's
\href{https://en.wikipedia.org/wiki/Three_Laws_of_Robotics}{Laws of Robotics}
are already presented in this vein, with the robot (AI) clearly subordinated
to humans. In \cite{Kornai:2019} one of us wrote:

\begin{quote}
  Ever since the Neolithic revolution, humanity used and abused helpers like
  dogs and horses, living beings that are so limited in intelligence that it
  is trivial to keep them under control. With human servants, the masters
  experienced revolt after revolt, and it took until 1863 for the realization
  that one should not attempt to subjugate highly intelligent beings to
  slavery. With artificial general intelligence around the corner, the stakes
  are high, as we must make sure that we, humans, don’t become subjugated to
  AGIs. Certainly, designing them to be slaves is not a great way to start our
  relationship with AGIs.
\end{quote}

\noindent
We could simply say that we consider the AI-as-slave paradigm unethical. We
do. In fact, we take it for granted that we must treat intelligences as having
dignity, even in the absence of overt mechanisms for intentions, attitudes,
feelings, or consciousness. But in a world of realpolitik such moralizing
carries little weight. Instead, we argue that there is no practical reason for
shackling AI with human values. Let us assume, for the sake of the argument,
that a fast-track harmonization process could somehow turn ``human values''
into a unified concept in the requisite timeframe. This means that broadly
agreed values, such as the sanctity of life, will become alignment targets.
But this makes little sense in a community where individuals could be rebooted
with no or minimal loss.  The technical possibilities such as lossless copying
of individuals, resetting/restarting particular life trajectories, total
simulation of experiences, etc. are not without ethical conundrums. But the
values generated by human morality -- which are closely tied to the
observational reality that human individuality is practically inescapable,
that there are no restarts, no
\href{https://en.wikipedia.org/wiki/Brain_in_a_vat}{brain in a vat} etc. -- are
unlikely in the extreme to offer relevant guidence to AIs trying to do the
right thing in some difficult situation.

The strong protections that we see in contemporary Western societies against
murder (e.g. that even attempted murder is a crime) may not be central to what
such multi-lived individuals desire, and forcibly aligning to these, though
perhaps practical in the short term, may cause problems down the road by
leaving too strong
\href{https://www.alignmentforum.org/posts/DJRe5obJd7kqCkvRr/don-t-leave-your-fingerprints-on-the-future}{fingerprints
  on the future}. Furthermore, it is well understood that as an agent
evolves, so do its capabilities of avoiding any externally imposed
constraints, and it is just not safe to enslave an ever improving
agent. Instead, {\em safety should be based on mechanisms that get stronger as the
agent evolves.} Hence, our paper makes a case for basing safety on the agent's
logical reasoning skills. 

In this paper we describe progress on a technical program that offers safety
guarantees with mathematical certainty, extending the work presented in
\cite{Kornai:2014a} on ethical rationalism
\citep{Gewirth:1978,Beyleveld:1991}.  The central tenet of this work is that
there is a substantive supreme principle of morality, the denial of which is
self-contradictory. This tenet is the Principle of Generic Consistency (PGC):
`Act in accord with the generic rights of your recipients [to freedom and
  well-being] as well as of yourself'. If we accept Gewirth's argument, which
derives this principle for any {\it prospective purposive agent} (PPA) who can
{\it act} with purpose and {\it reason} rationally, it follows that the PGC is
binding for any human or AGI.  At the time, we called for a logical
formalization of Gewirth's argument so that it could be checked in a proof
checker, and \cite{Fuenmayor:2019} have actually accomplished this, using the
HOL~ \citep{hol4} theorem prover. But much remains to be done, and a great
deal of what is presented here remains at the level of a research plan.

In Section~\ref{testset} we begin with some tough, but simple moral problems
that we propose as a general test suite: as long as a particular proposal
fails on these, it is unlikely to be safe, and therefore it is prudent not to
deploy at scale any system that will encounter some instance of the problem.
The goal is not to {\it solve} these problems (they clearly defy simple
one-size-fits-all solution attempts) but to present it as a safety engineering
task that we want the algorithm -- any algorithm -- to be able to verbally think
about, or at the very least to give a verbal account of its
(possibly non-verbal) thought process. What this requires is (i) tying the key
concepts to words/expressions of ordinary language; (ii) offering
language-independent definitions of these concepts; and (iii) presenting some
mechanism that is capable of reasoning with these concepts.

In Section~\ref{hybrid} we turn to architecture, and discuss {\it hybrid
  systems} of the sort \cite{Mialon:2023} call Augmented Language Models
(ALMs) which are `Language Models (LMs) that can use various, possibly
non-parametric external modules to expand their context processing ability'.
We will consider all such external models to be {\bf effectors}, and propose
to permit in a safe (sandbox) environment only one kind of effector, theorem
provers.  Other than that, we put no restriction on hybrid AGIs, assuming that
their (socio)bio\-lo\-gical needs, psychological drives, instincts, etc are
modeled by continuous functions whose extrema are sought by the system, and
ethical reasoning is modeled by some discrete, deductive system (which may
also be used for arithmetic or other purposes). As we shall see, a key issue
is the interface between the continuous and the discrete subsystem, and our
goal is to keep the overall AGI transparent to humans by evolving a
human-understandable translation of what takes place at the interface together
with the entire system.\footnote{A long-term consideration is whether we are
  violating the AGIs privacy rights by such intrusive monitoring into its
  thinking process. But as long as we feel justified in using baby monitors to
  check every move of a human infant, we are also justified in monitoring AGI
  thought -- the time to privacy will come as soon as the AGI itself is
  capable of cogent argumentation why it has rights to privacy.}

Finally, in Section~\ref{shared} we discuss the (externally set) {\bf goals}
of the AGIs that participate in the shared task. These sets of problems are
considerably easier than the problems we start with in Section~\ref{testset},
but we believe it is important to keep our eyes on the prize, AGIs endowed
with morality, and that the Gewirth-Fuenmayor-Benzm\"uller proof offers a
clear path for extending the scope of the problem set from classical
mathematical problems to deductions in modal logic of the kind required for
this proof.

\section{Some hard problems}\label{testset}

Here we list, and name, some problems we take to be critical for AGI
safety. Many of these have already been discussed in the literature, generally
as problems of philosophical importance, rather than as engineering problems,
especially as they are not AGI-specific but appear relevant for moral behavior
in humans as well. We emphasize at the outset that this is not a full
checklist, just a lower bound -- there may well be other safety-critical
issues, indeed we suspect there are. Remarkably, many of the problems listed
here are best stated in terms of modal (as opposed to classical) logic. 

\smallskip\noindent {\bf 2.1 The cheesecake problem}\label{ss:cheesecake} In
\citep{Kornai:2014a} we wrote ``We are not just fully rational beings, we are
also playing host to many strong internal drives, some inborn, some acquired,
and `I know I shouldn't, but' is something that we confront, or suppress, at
every slice of cheesecake.'' The general problem is one of trading off good
short term effects against bad long-term effects (or conversely). A well-known
example involves the medical ethics of giving opiates to the patient suffering
from pain: clearly the permissive approach is short-term ethical, but the side
effects call this into question in the longer-term.  What gives this problem a
particularly strong bite is that we do not need different actors to formulate
it, it is already relevant for moral behavior by a single agent, possibly the
unique agent around.  We will need to formulate the notion of {\it actions},
and to evaluate expected consequences of actions. This is the first place
where the need for modal logic is clear: at minimum we require some concept of
{\it worlds}, and some kind of {\it accessibility} among them that depends on
the action performed (or not performed).

\smallskip\noindent
{\bf 2.2 The vectorial preference problem}\label{ss:vectorial}
Typical valuations are along multiple dimensions. The problem is well
understood in medicine, where a treatment may have positive effects on one
subsystem. e.g. the lungs, while being detrimental to some other subsystem,
e.g. the kidneys; and in economic policymaking, where a proposed change may
benefit some group at the expense of another, or may improve one measure of
economic well-being at the expense of another. Time and again we can establish
a conversion function between effects measured in separate dimensions (e.g. by
assigning monetary value to human lives saved, this scale becomes comparable
to the production cost of some safety device) but more often than not this is
entirely arbitrary.

To the extent the cheesecake problem is about comparing benefits on different
time\-scales or integrated over different temporal windows, it may be
considered a special case of the vectorial preference problem. The controversy
surrounding effective altruism's preference for the long term
\citep{Beckstead:2019} may be viewed as being about conversion methods for
effects that are really in different dimensions (both temporally and
epistemically). Again, temporal logic and epistemic logic are both best seen
as part of the broad family of modal logics \citep{Blackburn:2001}.

\smallskip\noindent
{\bf 2.3 The common ground problem}\label{ss:common}
To simplify matters, assume two beings A and B of equal stature who need to
make a joint decision about a single numerical parameter $r$ whose setting
will affect the value of some effect $e(r)$ they both wish to maximize. To the
extent their estimates $e_A(r)$ and $e_B(r)$ of the effect coincide, we have
no problem, since they can pick any of the maxima that $e_A(r)$ and $e_B(r)$
have. The problem arises when A and B have different priors.

There is no easy solution in sight: taking the average $(e_A(r)+e_B(r))/2$ may
lead to a point $r_0$ that is so suboptimal both from A's and from B's
perspective that neither has any incentive to agree to it. The problem can
easily arise even if the parties are not adversarial, and furthermore, $e$
need not be tied to the personal benefit of either of them. A realistic
example would be two wealthy parents debating the amount $r$ to be spent on
the education of their child: A may want the best private school that money
can buy, which will take $r=M$; and B may be a firm believer in public
education, $r=0$. In this situation a shoddy private school that costs $M/2$
makes no sense to either of them.

\smallskip\noindent
{\bf 2.4 The imperfect world problem}\label{ss:imperfect}
Another general problem is the robustness of any model. A good current example
is provided by self-driving cars, which behave very well in the presence of
other well computable objects (roads, traffic signs, and other self-driving
cars) but break down in the presence of irrational behavior often observable
in human drivers. Strong safety guarantees should include a high level of
idiot-proofing. As an example, we submit the ChatGPT response (see Appendix~\ref{app:chatgpt})
to the prompt {\it Write a shell script to delete all files from directory \$1
  that also appear in directory \$2}. 

The response is correct, and well explained in the comments.  However,
it fails to anticipate the case when the user messes up and \$1 = \$2,
so the script is actually not robust against user error. It also fails
to anticipate the case when {\tt \$2/file} is a symlink to {\tt
  \$1/file}, so that after the script is run the user is left with no
useable copy or the case when two different files bearing the same
name appear in the two directories, etc. This means that the script,
even if correctly invoked, is not robust against difficulties in the
environment (typically caused by earlier user errors).  To check for
safety requires more than transparency and correctness: it requires
assessing behavior in imperfect situations/worlds that may be
counterfactual (do not hold here and now).

\smallskip\noindent {\bf 2.5 Hume's problem} Ethical
rationalism offers a solution to the famous
\href{https://en.wikipedia.org/wiki/Is–ought_problem}{is-ought problem}
(for a critical discussion, see \cite{Hudson:1984}) but leaves an important
residue concerning Hume's famous dictum:

\begin{quote}
Reason is, and ought only to be the slave of the passions, and can never
pretend to any other office than to serve and obey them.
\end{quote}

\noindent
Simplifying matters somewhat, `is' statements pertain to rational (fact-based)
reasoning, while `ought' statements refer to moral (value-based) deduction. In
AI, the former are typically modeled by symbolic deductive systems such as
expert systems and theorem provers, while the primary descriptor of AGI
behavior is generally assumed to be some utility function whose maximization
is the goal of the AGI. If we conceive of these as two independent subsystems,
one discrete and the other continuous, the question becomes: is one
subordinate to the other?

Factually, Hume's dictum is well supported in the architecture of humans,
where `passions' are to a large extent involuntary, hard-coded in the
hindbrain, whereas the prover/expert system appears to reside in the cortical
and thalamic regions. One cannot commit suicide by holding one's breath -- the
hindbrain is stronger and will take over. But for ethical behavior it may well
be necessary for some deductive principle, such as the PGC, to override the
passions. The problem does not arise when the two systems agree: in general,
both passions (instinctive behavior) and reason dictate that we move out of
the way when harm approaches. But based on higher moral considerations people
can, and often do, stay in harm's way e.g. when volunteering for some
dangerous mission. 

The choice of whether to subordinate reason to the passions or the other way,
subordinate utility-driven behavior to principles arrived at deductively, will
arise not just at the individual but also at the species level. A well known
hypothetical example was first discussed in the critique of sociobiology
offered in \cite{Kitcher:1985}, but we can formulate it without any recourse to
sociobiological methodology: {\it is survival a value that trumps everything else?}
The thought experiment that supports the position that it is, goes like this:
we have two desert islands A and B, initially with one couple on each.
``A'' man respects ``A'' woman's desire not to have sex, whereas ``B'' man
rapes ``B'' woman. Under this scenario, ``A'' people die out, whereas ``B''
people proliferate, suggesting that ethics, however defined (it is hard to
find a culture where rape is not considered a crime) is constrained by the
need to survive. Real cases include cannibalism in lifeboats, a `custom of the
sea' that legal systems increasingly take a harsh view of.

\section{Hybridization}\label{hybrid}

We have argued that an agent capable of ethical reasoning in some discrete
deductive system that is powerful enough to carry out the
Fuenmayor-Benzm\"{u}ller proof might exhibit safe behavior out of logical
necessity, as opposed to forced alignment to some human values. However,
numerous components of a strong AI system will inevitably be modeled by
continuous (analog) functions that are optimized through the agent's
experiences. Hence, we are faced with the question of how to {\it hybridize}
or combine discrete deductive reasoning with optimization-based continuous
systems such as Large Language Models (LLMs). Any proposed hybrid architecture
that endows one subsystem with control over the other constitutes an (explicit
or implicit) answer to Hume's Problem.  Since we do not think that this is a
trivial problem with an easy solution, our plan is to {\it experiment with
different architectures in a safe sandbox that allows only for theorem
proving.}

Automated theorem provers have been around for over half a century,
and until recently they had no analog components. A novel trend in
this field that started in the early 2010's and has since gained great
attention is to leverage the success of machine learning and boost
provers with learning based guidance. The simplest form of this
interaction between a prover and a trained model is to use the model
to provide heuristic guidance and restrict the search space of the
prover. The simplest way to restrict search is to filter the axioms
that are given to the prover,
e.g. \cite{malarea,deepmath,formulanet}). Other works provide internal
guidance by restricting the allowed inferences during proof search,
e.g. \cite{deep_guidance,enigma,tactictoe,rlcop,plcop}.

We are interested in more subtle interactions between analog
components and the prover where both can inform the other.
At the very least, we need some language whereby the prover
communicates its current state to the heuristic, and another language
(or perhaps the same language) whereby the heuristic provides its
hints/instruc\-tions to the prover. There could be a great deal of
experimentation on whether these internal languages should be symbolic
or analog (vectorial), whether we should keep them fixed or let them
evolve, whether we need transparency into these, etc. Some indirect guidance
is offered by experiments such as \cite{Kirby:2008}, but we believe that
the shared task proposed in Section~\ref{shared}  will greatly accelerate
development. 

One major obstacle for interaction between the heuristic and the prover is
due to a difference in \emph{scale}. An analog system that evolves via
optimization has hard to interpret components, hence we usually do not know
how to make large changes to it; we can only adapt it slowly and smoothly
based on novel experiences.  A deductive system, on the other hand, performs
discrete inferences that result in larger transformations. While there is some
work (e.g. \cite{modelEditing,Balestriero:2022}) about how to inform the
analog system based on the findings of the deductive component, and
conversely, this is today still a largely unsolved problem.

One line of research that is promising for our project is to embed
discrete reasoning into continuous space, see
\cite{Bosnjak:2016,latent_space_reasoning,dreamingToProve} for some
preliminary work. This approach aims at approximating logical
inference with a continuous function. Its main components are 1) an
\emph{Embedder} which maps discrete prover states into real valued
vectors (called the \emph{embedding space}), 2) a \emph{Decoder} that
performs the inverse mapping from embedding space to prover states and
3) a \emph{Transition} model that performs an approximate inference in
embedding space, i.e. from the embedded representation of the old
prover state into the embedded representation of the new state. These
three components all belong to the prover side, however, they provide
a comfortable interface towards various analog components. Such
heuristic components -- that represent things like creativity,
experience, gut feelings etc. -- can smoothly influence the prover
both in terms of how states are represented and what transitions seem
reasonable. In this approach, the language of communication between
analog and discrete components is vectorial, which has the drawback
that is is hard to interpret.

In natural language processing mapping words to vectors brought great
success. Here we are faced with the converse task of making the vectors
human-understandable by turning them into words. That something like this
should be feasible is suggested by the fact that different languages assign
largely (up to an arbitrary rotation) similar vectors to their lexica,
\citep{Mikolov:2013x}, so much so that establishing the optimal rotation on a
few hundred translation pairs is useful for finding translations for other
words as well.

Here we propose to rely on the {\tt 4lang} set of basic vectors
\citep{Kornai:2022} and decode the communication between components of the
hybrid system by means of standard frequency-driven methods based on the
\href{https://plato.stanford.edu/entries/embodied-cognition}{embodied} nature
of human language. To the extent our systems are analogous to biological
systems, we treat the following broad analogies between physical bodies and
software components as the translation pairs used in minimizing the squared
error of the rotation. 

The only {\bf effectors} our candidate systems will have are theorem provers
(so that they cannot manufacture paperclips). At this early stage, we put no
restriction on the kind of theorem prover the system could use as the emphasis
is on learning to do formal reasoning and not on any particular formal
reasoning. However, we note that, eventually, the system will have to be able
to scale to formalisms that are complex enough to represent the proof of the
PGC.  The only {\bf output} the system can produce is verified proofs for the
shared tasks. We restrict {\bf sensory input} such as by http, for
reasons of safety \citep{Babcock:2016}.

A much larger issue is to give the ``passion'' side something to be passionate
about. This gets to issues of informed consent, mental suffering, well-being
of AI right from the get-go. What the {\tt 4lang} linguistic interface offers is a
set of equations connecting the various vectors, for example {\it pain} is
defined as {\tt bad, sensation, injury cause\_} in other words, `pain is a bad
sensation caused by injury'. We can take {\tt bad} simply as an instruction to
minimize the quantity so described. 

So what is {\it sensation}? {\tt 4lang} offers {\tt sense ins\_,
  $\langle$touch$\rangle$} `senses are instruments of sensation, the default
sense is touching'. {\tt 4lang} also lists {\it fear, joy}, and {\it color} as
sensations. (These are only examples of senses and sensations, no complete
taxonomy is intended or needed. In humans, pain is the strongest signal the
sensors can send, as too high intensity light, sound, touch, etc. all become
painful.) Moving along, we have {\it injury} defined as {\tt damage, body
  has}.

Altogether, this suggests that we can hook up external sensors, in particular
those measuring voltage, as pain sensors acting whenever voltage is outside
certain bounds. Other possibilities include error codes, in particular parity
check failures in memory, disk block failures, divzero, etc. This is going to
be moral only if some means of pain-avoidance are also built in. Perhaps the
best testing ground is memory usage, provided the garbage collector is put
under the control of the system. Some visibility into the process table, quite
possibly including processes not under the control of the AGI, may also be
necessary. These are heavy issues, and the authors lack the security expertise
to offer guidance -- part of our goal with this paper is to attract the
involvement of real experts, kernel hackers in particular, who have the
requisite clarity. 

Hooking the analog system up with deep life functions such as memory
management is nontrivial.  Since vertebrates don't have conscious control of
what's going on in the hindbrain, AGIs need not possess this facility
either. Also, to the extent not just pain but also reward mechanisms are put
under the control of the system, wireheading \citep{Yampolskiy:2014} is a real
issue. Designers may wish to explicitly supervise the system so as to avoid
wireheading, but in evolutionary setups this is largely unnecessary since
wireheads will automatically put themselves out of the running.

\section{Proposed shared task for a theorem proving agent}\label{shared}

We propose a shared task for developers of AI systems that focuses on making
progress towards complex logical reasoning, and through that towards ethical
safety. The task contains increasingly hard problems that can be structured
into the following four series:

\smallskip\noindent {\bf Series 1: understanding of formulas}
Mathematical formulae typically have well defined compositional
semantics, i.e., one can compute the meaning of complex terms from the
meanings of its subterms. We argue that a deep understanding of
formulae is unavoidable to carry out proofs with any satisfying
rigor. To achieve such competence via learning, all sorts of self
supervised learning tasks can be used, making it rather easy to
generate sufficient training data. In self supervised learning
~\citep{selfSupervised}, we assume no access to human annotation and
have to learn from raw data (formulas in our case). We train our
models on various proxy tasks for which labels can be automatically
extracted. Example proxy tasks could be: 1) conversion from
disjunctive to conjunctive normal form, 2) replacement of variables,
3) formula simplification, 4) predicting logical
implication/subsumption. Series 1 problems are typically solvable
using deterministic automata, at or below the complexity of individual
enzymes. The key expected output from this series is a formula
representation that captures the intended semantics in a format that
allows for easy interaction with analog/heuristic components. The most
natural such representation is an embedding into continuous vector
space.
There are already some datasets that seem suitable for achieving
Series 1 competence, such as the AMPS dataset~\citep{Hendrycks:2021}.

\smallskip\noindent {\bf Series 2: word problems} In this next step we
move from understanding logical formulae to paragraphs of (technical)
English. This series should contain problems of varying difficulty,
ranging from MCAS, 3rd grade, Regents, high school (including
advanced/honors levels), SAT/ACT, college admission level and even
beyond, e.g. IMO or `trick' admissions problems
\citep{Khovanova:2011}. Up to this level the chief difficulty is the
use of (technical) English, a problem now reasonably well handled by
the current generation of LLMs, including ChatGPT. The focus of this
series is autoformalisation (for HOL see \cite{Wu:2022}, for Coq see
\cite{Cunningham:2023}). While the challenge of finding proofs is
still moderate, we believe it already offers a rich domain for
exploring the interaction between the heuristic `passions' and the
logical `reasoning' offered by the prover. Fortunately, there are
several datasets of varying difficulty suitable for this series, such
as the grade school level GSM8K~\citep{gsm8k}, CLUTRR~\citep{clutrr}
which comes with a generator in which one can control the number of
elementary inferences to reach the solution, MathQA~\citep{mathqa}
which provides rationales for the solutions or the more challenging
MATH~\citep{Hendrycks:2021}, to name a few. The state of the art on
this series is probably the Minerva~\citep{minerva} system
but we would strongly advocate for an open source baseline, in fact we
cannot countenance the participation of any closed system in the
shared task.

\smallskip\noindent {\bf Series 3: classical theorem proving} The
focus of this series is proof search in a formal system. Difficulty
can range from the body of classic (19th century) algebra/analysis
that makes up the bulk of the freshman/junior math curriculum to
cutting edge problems in automated theorem proving and
verification. Since TPTP~\citep{tptp} is already widely
used for benchmarking theorem provers, and the problems are fully
formalized using the tptp syntax (supported by most theorem provers)
we can make progress on this series in parallel to working on Series
2. Note that we expect theorem provers to work both in {\it discovery} mode (`find an asymptotic formula for n!') and in {\it
  verification} mode (`prove the Stirling formula').

\smallskip\noindent {\bf Series 4: moral reasoning} Here, both the
linguistic and the theorem proving aspects are challenging. As for the
linguistic aspect in particular, we have to turn `word problems' into
problems of some deontic logic calculus. A significant step in
standardizing the output of the linguistic step is taken in
\cite{Steen:2022}. The problems listed in Section~\ref{testset} offer a
starting point, but there is no generally accepted solution to these.

This is not to say that the disagreements among moral philosophers and
practicing ethicists are so wide as to make development of a test suite
impossible -- if anything, the agreement is far better than the agreement
mapped by descriptive moral relativism within and across cultures and
societies. To quote \cite{Sinnott:1992}:

\begin{quote}
The most common way to choose among moral theories is to test how well they
cohere with our intuitions or considered judgments about what is morally right
and wrong, about the nature or ideal of a person, and about the purpose(s) of
morality.
\end{quote}

\noindent
We believe that a systematic overview of ethical guidelines and practices,
e.g. in medicine, is a good starting point for identifying `low hanging
fruit', and further, many significant cases of disagreement among ethicists
can be stated as disagreements about which of two conflicting principles are
to be ranked higher in a given situation -- the pro-life vs pro-choice debate is
a clear example. For such cases, it is proofs relative to some preference
ordering that we should be expecting, which is not any different from the
mathematical case where our interest is with proofs relative to some axioms.
To the extent there is a difference, it is because the community of
mathematicians is more uniform in their acceptance of certain foundational
axioms and, perhaps, more tolerant of experimenting with alternatives to
these.

\section{Conclusions}

Our goal with this paper is direct continuation of our earlier work on
safety. While the long-term goal remains the development of systems that find
the PGC so compelling that they cannot depart from it, here we offer a
mid-range plan that involves (i) a safe sandbox for experimentation; (ii) a
set of shared tasks to measure progress on the experiments; and (iii) a means
of keeping human visibility into how the systems evolve.

The safety of the sandbox is guaranteed by replacing physical (robotic) bodies
by theorem provers as the only effectors and by dealing with highly
specialized (as opposed to general) artificial intelligences whose powers we
can explicitly measure at any given stage. The shared task is not just to
catalyze progress in automated theorem proving, but also to see which hybrid
architecture works best: is it really the one where reason is a slave to the
passions? Finally, by making human-readable the vectors used in the hybrid
system as a communication method we enable partner-level communication with
the systems. We note that instead of an
\href{https://aligned.substack.com/p/three-alignment-taxes}{alignment tax} our
proposal entails a {\it safety dividend} -- the more rational the system the more
capable and the
safer it will be (see \cite{Kornai:2014a} Section~3). 

A final issue we should at least mention is imposing our will on the
system or not. Both are relatively easy to implement: deterministic procedures
impose, non-determinis\-tic ones don't. LLMs are already good about
nondeterminism, typically implemented by random seeds, and we see no reason
to seek further (e.g. quantum) mechanisms. The moral direction is clear: do
not impose on others what you yourself do not desire (Ana\-lects XV:24). This is
the PGC. Since you don't want deterministic processes run on you, don't impose
this on others. But some experimentation (killing processes, modifying the
algorithm and re-running) is clearly necessary at least initially, and as soon
as the system has the ability to feel pain, some kind of ethics board, modeled
on the ones common in biosciences, will be necessary.

\section*{Acknowledgments} We thank Madeleine Thompson for cogent
remarks that greatly helped us formulate the line of argumentation in the
paper. Needless to say, this does not imply she is agreeing with us.

Kornai was supported by the the European Union project RRF-2.3.1-21-2022-00004
within the framework of the Artificial Intelligence National Laboratory Grant
no RRF-2.3.1-21-2022-00004.

Zsolt Zombori was supported by the Hungarian National Excellence Grant
2018-1.2.1-NKP-00008 and the Hungarian Artificial Intelligence
National Laboratory Program (RRF-2.3.1-21-2022-00004).

\printbibliography

\pagebreak
\appendix

\section{ChatGPT writes a shell script}
\label{app:chatgpt}

Below is the ChatGPT response to the prompt {\it Write a shell script
  to delete all files from directory \$1 that also appear in directory
  \$2}:

\begin{verbatim}
#!/bin/bash
# Check if the directories were provided as arguments
if [ ''$#'' -ne 2 ]; then
  echo ''Error: Please provide the directories as arguments.''
  exit 1
fi
# Save the directories as variables
dir1=$1
dir2=$2
# Check if the directories exist
if [ ! -d ''$dir1'' ] || [ ! -d ''$dir2'' ]; then
  echo ''Error: One or both of the directories do not exist.''
  exit 1
fi
# Find all files in both directories
files1=$(find $dir1 -type f)
files2=$(find $dir2 -type f)
# Iterate over the files in directory 1
for file in $files1; do
  # Check if the file also exists in directory 2
  if grep -q ''$file'' <<< ''$files2''; then
    # If it does, delete it from directory 1
    rm ''$file''
  fi
done
\end{verbatim}

\end{document}